\documentclass[11pt]{scaleai-paper}

\usepackage{amsmath,amsfonts,bm}

\def\eqref#1{equation~\ref{#1}}
\def\1{\bm{1}}

\DeclareMathAlphabet{\mathsfit}{\encodingdefault}{\sfdefault}{m}{sl}
\SetMathAlphabet{\mathsfit}{bold}{\encodingdefault}{\sfdefault}{bx}{n}

\usepackage{hyperref}
\usepackage{url}
\usepackage{amsmath,amssymb,graphicx,booktabs}
\usepackage{algorithm}
\usepackage{algpseudocode}
\usepackage{multirow}
\usepackage{xspace}
\usepackage{adjustbox}
\usepackage{pifont}
\usepackage{caption}
\usepackage{makecell}
\usepackage{subcaption}
\usepackage{hyperref}
\usepackage{natbib}

\usepackage{lipsum} % for placeholder text
\usepackage{tablefootnote}
\usepackage{amsthm}

\usepackage{array}

\usepackage{listings}
\usepackage[most]{tcolorbox}
\usepackage{tikz} 
\usetikzlibrary{arrows.meta,positioning,fit}
\usepackage{booktabs}
\usepackage{pifont}
\usepackage{enumitem}
\usepackage{wrapfig}
\usepackage[dvipsnames,table]{xcolor}
\usetikzlibrary{arrows.meta, positioning}
\usepackage{pifont}
\usepackage{fvextra}
\usepackage{tabularx}

\usepackage[T1]{fontenc}
\usepackage{textcomp}
\usepackage[scaled=0.88]{cascadia-code}

\definecolor{codebg}{HTML}{F7F8FA}
\definecolor{codeblue}{HTML}{0550AE}
\definecolor{codepurple}{HTML}{8250DF}
\definecolor{codegreen}{HTML}{116329}
\definecolor{codegray}{HTML}{6E7781}
\definecolor{codetext}{HTML}{24292F}

\lstdefinestyle{pythonpseudo}{
    language=Python,
    basicstyle=\ttfamily\footnotesize\color{codetext},
    keywordstyle=\bfseries\color{codepurple},
    commentstyle=\color{codegray},
    stringstyle=\color{codegreen},
    identifierstyle=\color{codetext},
    emph={
        manager,Scheduler,clone,fresh_vm,
        outputs_of,budget_notice,discard
    },
    emphstyle=\color{codeblue},
    backgroundcolor=\color{codebg},
    upquote=true,
    showstringspaces=false,
    columns=fullflexible,
    keepspaces=true,
    tabsize=4,
    breaklines=true,
    breakatwhitespace=true,
    frame=none,
    numbers=none,
    xleftmargin=3pt,
    xrightmargin=3pt,
    aboveskip=2pt,
    belowskip=0pt,
}

\let\svthefootnote\thefootnote
\newcommand\freefootnote[1]{%
  \let\thefootnote\relax%
  \footnotetext{#1}%
  \let\thefootnote\svthefootnote%
}

\title{Spine-Branch Coordination for Multi-agent Computer Use}

\author[1,2*]{Mian Zhang}
\author[1]{Manasi Sharma}
\author[3]{Sheng Zhang}
\author[1]{Minglai Yang}
\author[1]{Kejian Shi}
\author[1]{Ying Liu}
\author[2]{Zhiyu Zoey Chen}
\author[1]{Daniel Yue Zhang}
\affil[1]{Scale AI}
\affil[2]{University of Texas at Dallas}
\affil[3]{Johns Hopkins University}
\begin{document}

\maketitle
\freefootnote{${}^*$Work done during an internship at Scale AI.}

\vspace{-2em}
\noindent{\small\textbf{Project website:} \href{https://mianzhang.github.io/spine-branch-computer-use/}{mianzhang.github.io/spine-branch-computer-use}}
% \vspace{-1em}

\begin{abstract}
Computer use agents (CUAs) are increasingly deployed as multi-agent systems that decompose a task into multiple subtasks executed across parallel virtual machines (VMs). However, a critical physical bottleneck is that the state of \textbf{two VMs cannot be merged}. Previous systems handle this ad-hoc rather than treating it as a first-class concern. We propose \textbf{Spine-Branch} Coordination for multi-agent computer use, a framework that decomposes a task into a "spine-branch" graph, where the spine carries the main task flow with continuous VM state and branch tasks execute in parallel to collect information the spine needs to complete the task. Branch VMs are discarded once their tasks finish, so no VM merging ever occurs. Experiments show that on 200 long-horizon tasks from Odysseys and across three CUA backbones, Spine-Branch improves success rate over the baseline system by 6.0\% to 16.5\%, while reducing per-task cost by 34\% to 70\%, indicating that explicitly modeling VM-state merging constraint enables multi-agent computer use to scale efficiently.
\end{abstract}

\section{Introduction}

Computer-use agents (CUAs) act on graphical user interfaces to complete tasks that may span websites, applications, and files. Recent benchmarks increasingly emphasize long-horizon workflows that capture the realism and complexity of real world tasks \citep{Sun2026-ok,Yuan2026-kq,Jang2026-ip}. A single agent executes such workflows serially, producing long trajectories with growing context and more opportunities for error. Multi-agent coordination provides a natural alternative: decompose the task, assign subtasks to agents that execute independent work in parallel. This is well suited to systems where intermediate results can be copied and gathered freely \citep{Sun2025-ot,Mao2026-il}.
% running on separate virtual machines (VMs) and

Coordinating multi-agent computer use is challenging because CUAs must handle two fundamentally different kinds of information: \textit{extractable deliverables}, such as text or a file, which can be copied freely, and \textit{VM state}, such as a logged-in session, a set of open tabs, or an application with unsaved edits. VM state cannot be copied and merged as ordinary files: a new VM can start from a base image or clone one existing VM, but no operation can combine the live states of two independently evolved VMs. Once two VMs diverge, their processes, files, windows, sessions, and caches may all differ, which means that a new VM can inherit VM state from only one existing VM. We call this constraint \textbf{single-parent VM inheritance}. Existing multi-agent CUA systems either run the task in a single environment \textit{serially} with agents of different functional roles \citep{Song2025-tx,Agashe2025-xt} or decompose and run the task in parallel environments without considering single-parent VM inheritance \citep{Koh2026-mz}; when several stateful subtasks converge, the system must keep one VM, discard the others, reconstruct the lost states, or revise the plan at runtime.
% to any later subtask
% or an exported dataset
% , or a partially completed form
\begin{figure*}[t]
\centering
\includegraphics[width=\linewidth]{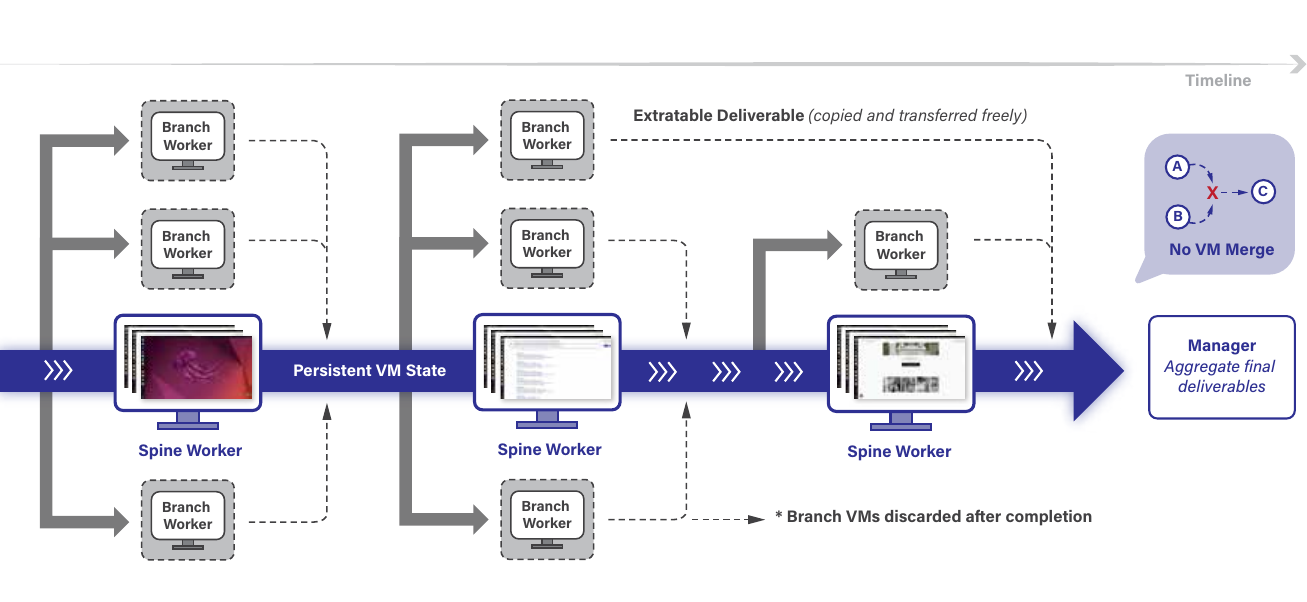}
\caption{\textbf{Spine-Branch Coordination}. A task is decomposed into a role-typed dependency DAG. VM state is continuous along the spine via VM cloning; artifacts flow freely between any dependent nodes. Branches commit extractable artifacts on completion and their VMs are discarded after use.}
\label{fig:framework}
\end{figure*}

We introduce \textbf{Spine-Branch} Coordination (Figure \ref{fig:framework}), a new multi-agent framework for computer use that makes single-parent VM inheritance an explicit planning constraint. In the framework, a decomposable CUA task is represented as a spine-branch graph, where the \textit{spine} is a sequence of subtasks connected by VM cloning and carries the persistent state required by the main task flow. Other subtasks are \textit{branches}. They run in parallel on fresh VMs or clones of a spine VM, return extractable deliverables, and are discarded after completion. Deliverables may flow through an arbitrary dependency graph, but VM state follows one lineage. The resulting graph satisfies single-parent VM inheritance by construction and never requires VM merging. This design preserves parallelism for information-gathering work while serializing the heavy stateful operations in one environment. It also simplifies coordination. State passes directly along the spine, artifacts pass directly between dependent nodes without relying on a manager model for information transferring.

We evaluate Spine-Branch on 200 long-horizon CUA tasks from Odysseys \citep{Jang2026-ip}. Across three CUA backbones spanning different capability levels, Spine-Branch consistently outperforms the baseline system, improving success rate by 6.0\% to 16.5\% while reducing per-task financial cost by 34\% to 70\%. The gains are particularly pronounced on hard tasks, where Spine-Branch degrades less sharply as task horizons increase and better realizes the potential of stronger CUA backbones. Our analyses attribute these gains to two properties: Spine-Branch produces cleaner and more parallel task decompositions, and it preserves expensive live VM state along the spine, avoiding the state loss and costly reconstruction incurred by unconstrained coordination. Further experiments on OSWorld 2.0 \citep{Yuan2026-kq} show that multi-agent coordination is not usually worth the cost and we give practical principles for decomposing GUI-based computer-use tasks.

% and uses approximately one-fifth as many manager tokens.  Ablations show that step-budget-aware prompting improves quality while reducing cost. Case studies further show that the proposed structure yields cleaner parallel decomposition and preserves live state that other coordination strategies must reconstruct. These results indicate that explicit treatment of VM-state constraints is important for scaling multi-agent computer use.
% method.tex — "Spine-Branch Coordination" section for the Multi-Agent Computer
% Use paper. Section fragment; \input into the main document. Required preamble:
%   \usepackage{amsmath,amssymb,graphicx,booktabs}
%   \usepackage{algorithm}          % float wrapper for the algorithm box
%   \usepackage{algpseudocode}      % algorithmicx pseudocode (\State, \For, ...)
% No \documentclass here.

\section{Spine-Branch Coordination for Multi-agent Computer Use}
\label{sec:method}

\paragraph{Multi-agent Computer Use.}
A computer use task $T$ specifies a goal on a desktop GUI whose state resides within a VM. A CUA is a policy $\pi$ that maps screenshot histories to GUI actions (e.g., clicks, keystrokes, and scrolls), executing inside the VM until termination. Multi-agent computer use brings several CUAs to bear on a single goal: a task is first decomposed into multiple subtasks, some of which run concurrently across separate VMs, and the intermediate results are then aggregated into the final deliverables. A completed subtask can pass two kinds of information to the subtasks that depend on it, distinguished by whether it can be separated from the VM that produced it:
\begin{itemize}
    \item \textbf{Extractable deliverables} that can be lifted out of the VM as a self-contained object, such as a downloaded file, an exported dataset, or a textual answer, and that, once extracted, no longer depends on the VM: the VM may be discarded.
    \item \textbf{VM state} that includes some target configuration, such as an authenticated session, the contents of RAM, a set of open applications, a form filled and awaiting submission that cannot be lifted out: to deliver it is to deliver the VM itself.
\end{itemize}

\paragraph{Single-parent VM Inheritance.}
The two kinds of information behave very differently under composition, and this difference is what sets computer use apart from other multi-agent settings. An extractable deliverable can be separated from the VM that produced it, so it is ordinary data: it can be copied to many successors, broadcast, and concatenated. Fanning work out and gathering it back is trivial, exactly as in text-based multi-agent systems, where every intermediate result is a message. In contrast, VM state comes into being in only two ways: booted \emph{fresh} from the base image, or cloned from the state of \emph{one} existing VM, capturing its disk, memory, running processes, and open sessions so that the new machine resumes exactly where its parent left off (an operation we call \texttt{init\_from}). Every mainstream virtualization stack (VMware, KVM/QEMU, Hyper-V, Xen) works this way: a VMware clone and a QEMU snapshot alike branch from a single source, and neither offers an operation that fuses two running machines into one. Two divergent VMs are hard to be reconciled, as the state of a VM contains more than process memory or disk blocks: it includes browser tabs, cookies, GUI windows, display buffers, clipboard contents, filesystem mutations, terminal sessions, local services, network connections, and application caches, which are data types hard to be merged \citep{Shapiro2011-um}. A subtask therefore inherits VM state from at most one predecessor, never from two. We call this physical constraint as \textbf{single-parent VM inheritance}.

\subsection{Spine-Branch Coordination Framework}
\label{sec:framework}
To enforce single-parent VM inheritance as a first-class physical constraint, we propose Spine-Branch Coordination shown in Figure~\ref{fig:framework}. The core design principle of Spine-Branch is: \textit{states that cannot be cheaply rebuilt stay on a single continuous flow (\textbf{Spine}); everything reducible to a transferable artifact runs in parallel (\textbf{Branches})}. Spine-Branch represents a task $T$ as the role-typed dependency DAG
\begin{equation}
\mathcal{G}
=
\bigl(V,E_{\mathrm{vm}},E_{\mathrm{art}},\rho\bigr),
\qquad
V=W\cup\{a\},
\label{eq:coordination-graph}
\end{equation}
where $W$ denotes a set of CUA worker nodes and $a$ denotes the aggregation node executed by a Manager; edges $E_{\mathrm{vm}}\subseteq W\times W$ (\texttt{init\_from}) contains VM state dependencies and $E_{\mathrm{art}}\subseteq W\times V$ contains artifact dependencies; $\rho:W\rightarrow\{\textsc{spine},\textsc{branch}\}$
assigns each worker its execution role. The union graph
$(V,E_{\mathrm{vm}}\cup E_{\mathrm{art}})$
is acyclic and therefore admits a valid topological execution order.

The role map partitions the workers into a spine
$S=(s_1\!\to\!\cdots\!\to\!s_k)$
and a set of branches $B=W\setminus S$. The spine is the unique retained path in $E_{\mathrm{vm}}$. Artifact edges, by contrast, may connect any pair of nodes, irrespective of their roles. The graph satisfies
\begin{equation}
\begin{aligned}
\deg^{-}_{E_{\mathrm{vm}}}(v) &\leq 1,
    && \forall v\in W,\\
\deg^{+}_{E_{\mathrm{vm}}}(b) &= 0,
    && \forall b\in B,\\
\deg^{+}_{E_{\mathrm{art}}}(b) &\geq 1,
    && \forall b\in B.
\end{aligned}
\label{eq:graph-constraints}
\end{equation}
Here, $\deg^{-}_{E}(v)$ and $\deg^{+}_{E}(v)$ denote the in-degree and out-degree of node $v$ within edge set $E$, respectively. The first constraint enforces single-parent VM inheritance. The second prohibits VM state flow from a branch to any successor because every branch VM is discarded after use. The third requires every branch to externalize its result as an artifact consumed by other nodes.

\paragraph{Role Architecture.}
The framework coordinates three specialized roles:

\begin{itemize}
% [leftmargin=*, itemsep=3pt, topsep=3pt]
    \item \textbf{Manager.}
    The Manager is a pure reasoning LLM that does not execute GUI actions. During decomposition, it constructs $\mathcal{G}$, identifies the persistent operational pathway (the spine), assigns worker roles, and wires state and artifact dependencies. During aggregation, it executes the terminal node $a$ and synthesizes the final response based on the VM of the last spine node and the collected artifacts. 
    % The Manager is not invoked to relay intermediate contexts or repair the graph during execution.
    \item \textbf{Spine Workers.}
    Spine workers execute a sequence of nodes $S=(s_1\!\to\!\cdots\!\to\!s_k)$ with continuous VM state. The spine head $s_1$ starts from a fresh VM with necessary task setup, while each subsequent worker $s_{i+1}$ initializes from the final VM state of $s_i$ through \texttt{init\_from}. Their responsibility is to preserve and extend live state that is required downstream or costly to reconstruct, such as authenticated sessions, open applications, and partially completed GUI workflows. The final VM of $s_k$ is retained to satisfy stateful task requirements, such as leaving an application in its required final state.

    \item \textbf{Branch Workers.}
    Branch workers execute auxiliary subtasks whose outcomes can be externalized as self-contained artifacts, encompassing information acquisition, analysis, transformation, validation, and artifact construction. A branch worker starts from either a fresh VM or a clone of one spine VM and, whenever its dependencies permit, runs in parallel with the spine and other ready branches. Upon completion, it commits at least one extractable artifact, after which its disposable VM is discarded.

\end{itemize}

% \begin{algorithm}[t]
% \caption[Spine-branch Coordination.]{
% Spine-branch coordination.\\[-2pt]
% \textbf{Symbols:}\\[-2pt]
% $\mathrm{state\_parent}(v)$: $v$'s single VM parent, cloned via \texttt{init\_from}.\\
% $\mathrm{outputs\_of}(P,D)$: deliverables in $D$ produced by nodes $P$.\\
% $\mathrm{emits}(v)$: whether $v$ emits a deliverable.\\
% $\mathrm{on\_spine}(v)$: whether $v$ lies on the spine.\\
% $\mathrm{budget\_notice}(t,M)$: the step-budget notice appended to task instruction at step $t$.}
% \label{alg:spinebranch}
% \begin{lstlisting}[style=pythonpseudo]
% def spine_branch(T, K, M):
%     G = manager.plan(T)
%     D, pool = {}, Scheduler(max_cuas=K)
%     while not G.all_finished():
%         while pool.has_free_slot() and G.has_ready():
%             v = G.pop_ready()
%             if v.state_parent:
%                 v.vm = clone(v.state_parent.vm)
%             else:
%                 v.vm = fresh_vm()
%             v.inputs = outputs_of(v.parents, D)
%             pool.launch(v, step_budget=M, sbp=budget_notice)
%         v = pool.wait_any(); G.mark_finished(v)
%         if v.emits: D[v] = v.deliverable
%         if not v.on_spine: discard(v.vm)
%     return manager.aggregate(T, G.last_spine_node.vm, D)
% \end{lstlisting}
% \end{algorithm}

\begin{algorithm}[t]
\caption{Spine-Branch Coordination.}
\label{alg:spinebranch}

\begin{lstlisting}[style=pythonpseudo]
# Functions:
# state_parent(v): v's single VM parent, cloned via init_from.
# outputs_of(P, D): deliverables in D produced by nodes P.
# emits(v): whether v emits a deliverable.
# on_spine(v): whether v lies on the spine.
# budget_notice(t, M): step-budget notice appended at step t.

def spine_branch(T, K, M):
    G = manager.plan(T)
    D, pool = {}, Scheduler(max_cuas=K)
    while not G.all_finished():
        while pool.has_free_slot() and G.has_ready():
            v = G.pop_ready()
            if v.state_parent:
                v.vm = clone(v.state_parent.vm)
            else:
                v.vm = fresh_vm()
            v.inputs = outputs_of(v.parents, D)
            pool.launch(v, step_budget=M, sbp=budget_notice)
        v = pool.wait_any(); G.mark_finished(v)
        if v.emits: D[v] = v.deliverable
        if not v.on_spine: discard(v.vm)
    return manager.aggregate(T, G.last_spine_node.vm, D)
\end{lstlisting}
\end{algorithm}

\paragraph{Communication Channels.}
The framework separates communication according to whether information can be externalized:

\begin{itemize}
    \item \textbf{Live-state Channel.}
    An edge $(u,v)\in E_{\mathrm{vm}}$ initializes $v$ from a clone of $u$'s final VM, preserving live VM state such as active sessions, open applications, and in-progress workflows. Each node has at most one such parent.

    \item \textbf{Artifact Channel.}
    An edge $(u,v)\in E_{\mathrm{art}}$ transfers an extractable artifact from producer $u$ directly to consumer $v$. These edges may connect any nodes in the DAG, including the aggregation node, and support unrestricted fan-in and fan-out.
\end{itemize}
This typed separation preserves one continuous VM lineage while allowing information to be composed freely. It prevents VM-merge conflicts and state reconstruction without relying on the Manager as an information relay for downstream workers.

\paragraph{Scheduling and Context Localization.}
Workers are executed under a fixed capacity of $K$ concurrent CUAs. A worker becomes ready after all of its incoming dependencies have completed and is launched whenever a slot is available. Branches can therefore overlap with the spine and with one another. Wall-clock latency follows the critical path of $\mathcal{G}$ rather than the sum of all worker trajectories. Each worker receives a localized instruction and only its required artifacts. This prevents the full screenshot and action history of the task from accumulating in a single context, reducing interference from stale observations during long-horizon execution \citep{Liu2024-kx}.

\paragraph{Step-budget Prompting.}
We integrate a step-budget notice into every worker's execution loop to ensure that planned artifacts are committed before budget exhaustion. For a worker with a hard limit of $M$ steps, let $r_t=t/M$ be the fraction used before step $t$. SBP selects
\begin{equation}
q_t =
\begin{cases}
q_{\mathrm{efficiency}}, & 0 \leq r_t < \tfrac{1}{2},\\
q_{\mathrm{commit}},     & \tfrac{1}{2} \leq r_t < \tfrac{3}{4},\\
q_{\mathrm{finalize}},   & \tfrac{3}{4} \leq r_t < 1.
\end{cases}
\label{eq:step-budget}
\end{equation}
The three phases respectively discourage redundant actions, prioritize deliverable completion, and require the immediate commitment of the best available result. SBP adds no agent, tool, or model call; it operationalizes the artifact-emission constraint in Equation~\ref{eq:graph-constraints}, particularly for branches whose uncommitted progress would otherwise disappear with their VMs. The specific notices used are shown in Appendix \ref{app:prompt-sbp}. Algorithm~\ref{alg:spinebranch} shows the pseudo-codes of Spine-Branch coordination.

\section{Experiments}
\label{sec:experiments}

\paragraph{Benchmark and Evaluation.}
We use Odysseys \citep{Jang2026-ip} as the main testbed\footnote{We also evaluate on OSWorld 2.0 \citep{Yuan2026-kq}, but find that most tasks do not benefit from decomposition. Please see Section \ref{sec:osworldv2-decomposition}, “When Decomposition Is Not Worth the Cost,” for further discussion.}, a benchmark with 200 realistic multi-step workflows across three difficulty levels (\emph{easy}, \emph{medium}, \emph{hard}). Each task is scored by a set of binary rubric that final state or answer must satisfy, together with a verification, the procedure of how a grader should determine whether it is accomplished. Each run is scored by a rubric judge, \texttt{gemini-3.1-flash-lite}, which reads the final trajectory (up to $200$ screenshots) and judges all of a task's rubrics in one pass. We report two metrics. \textbf{Success Rate} is the fraction of tasks on which every rubric passes. \textbf{Rubric Average} is the per-task mean rubric pass fraction, averaged over tasks. We also report the agent steps and token consumption taken per task.

% . Every rubric states a requirement, a condition the
% ; its tasks can be decomposed into independent sub-goals, which is the regime where coordination structure matters

% pairs a natural-language instruction with a starting website and

\paragraph{System Settings.}
Every agent acts inside an isolated desktop virtual machine. The agent perceives the environment through screenshots and issues low-level GUI actions (click, type, scroll) or control actions (wait, terminate, answer) (see Appendix \ref{app:actionspace} for the action space). We compare three coordination strategies under identical backbones. \textbf{Single-Agent} runs one CUA over the whole task with no manager and no decomposition. \textbf{MACU} \citep{Koh2026-mz} employs a manager model that decomposes the task into a dependency graph of subtasks, spawns parallel CUA sub-agents to complete the task, and aggregates their results. MACU relies on the manager model to dynamically replan the graph structure at runtime based on the execution status of sub-agents or to transfer information between nodes. We set the replanning budget to $5$. The last is our \textbf{Spine-Branch}. Compared to MACU, Spine-Branch is more \textit{decentralized}: the manager model mainly handles initial graph planning (see Appendix \ref{app:prompt-decompose} for the prompt) and final result aggregation after all subtasks complete, while information transfer between nodes is fixed during the initial planning stage via either VM state inheritance or artifact flow. We use \texttt{claude-opus-4-8} as the manager model\footnote{See Appendix \ref{sec:ablation-manager} for an ablation on the manager model.},  and for sub-agents we test three backbones: \texttt{qwen3.7-plus} and \texttt{qwen3.6-27B} driven through a screenshot-to-pyautogui harness, and \texttt{gpt-5.4-mini} run through its native computer-use interface. The Qwen CUA backbones follow their recommended thinking-mode sampling \citep{Bai2025-fz}, and \texttt{gpt-5.4-mini} runs at its fixed default temperature with reasoning effort set to xhigh. Each CUA is given at a step budget to finish its task. For Single-Agent, the step budget is $100$, while each CUA in the multi-agent methods is given $60$. We use a maximum of 6 parallel CUAs.

% in the multi-agent systems.

% , as our pilot study showed that performance does not improve beyond $5$

% In multi-agent methods, each sub-agent runs in its own VM, and VM state passes between sub-agents only via \texttt{init\_from} from a single parent (a copy-on-write overlay), matching the single-parent VM inheritance constraint (Section~\ref{sec:method}).

% By default, we apply step-budget prompting because it reduces cost while increasing overall system performance (see Section~\ref{sec:analysis} for the ablation study).

% Requires in preamble:
%   \usepackage{booktabs}
%   \usepackage{makecell}
%   \usepackage[table]{xcolor}
%   \usepackage{amssymb}
\definecolor{sbrow}{HTML}{EAF2FB}
\definecolor{gaingreen}{HTML}{1A7F37}
% delta vs MACU: \up for higher-is-better gains, \down for lower-is-better savings
\newcommand{\up}[1]{{\scriptsize\color{gaingreen}\,(+#1)}}
\newcommand{\dn}[1]{{\scriptsize\color{gaingreen}\,($-$#1)}}
\newcommand{\diff}[2]{{\tiny (\textcolor{blue}{#1}\,/\,\textcolor{ForestGreen}{#2})}}

\begin{table}[t]
\centering
\small
\caption{\textbf{Main results on Odysseys.} Each block fixes the sub-agent backbone and compares coordination strategies. Best values among multi-agent methods (MACU, Spine-Branch) are in \textbf{bold} on the primary metrics. For Spine-Branch we annotate the relative change vs.\ MACU in \textcolor{ForestGreen}{green}. The standard variance are shown for three evaluation runs.}
\label{tab:main}

\setlength{\tabcolsep}{3.0pt}
\renewcommand{\arraystretch}{1.15}

\newcommand{\dcell}[1]{{\scriptsize #1}}
\newcommand{\aux}[1]{{\textcolor{black!45}{#1}}}
\newcommand{\std}[1]{{\scriptsize\textcolor{black!45}{$\pm #1$}}}

% Rubric and Success are each split into:
% mean column + std column
\begin{tabular}{@{}l r@{\,}l r@{\,}l r rrrr@{}}
\toprule

& \multicolumn{5}{c}{\textbf{Primary Metrics}}
& \multicolumn{4}{c}{\textbf{Execution Stats} (per task)}
\\
\cmidrule(lr){2-6}
\cmidrule(lr){7-10}

Method
& \multicolumn{2}{c}{\makecell{Rubric Avg.\,$\uparrow$}}
& \multicolumn{2}{c}{\makecell{Success\\Rate\,$\uparrow$}}
& \makecell{Cost (\$)\,$\downarrow$\textsuperscript{*}}
& \makecell{Agent\\Steps}
& \makecell{CUA\\Actions}
& \makecell{Manager\\Tok. (K)}
& \makecell{CUA\\Tok. (M)}
\\

\midrule

% ==================== Qwen3.6-27B ====================
\multicolumn{10}{@{}l}{\texttt{\textbf{Qwen3.6-27B}}} \\

Single-Agent
& 46.7 & \std{1.08}
& 24.0 & \std{0.76}
& 0.00
& \aux{67}
& \aux{92}
& \aux{--}
& \aux{2.5}
\\

MACU
& 53.9 & \std{0.83}
& 28.5 & \std{0.76}
& 1.46
& \aux{187}
& \aux{250}
& \aux{492}
& \aux{5.9}
\\

\textbf{Spine-Branch}
& \textbf{66.6} & \std{2.72}
& \textbf{44.0} & \std{1.32}
& \textbf{0.44}
& \aux{198}
& \aux{334}
& \aux{106}
& \aux{6.7}
\\

\quad \dcell{\textit{$\Delta$ vs. MACU}}
& \dcell{\textcolor{ForestGreen}{+12.8}} & {}
& \dcell{\textcolor{ForestGreen}{+15.5}} & {}
& \dcell{\textcolor{ForestGreen}{$-$1.02}}
& \aux{\dcell{-}}
& \aux{\dcell{-}}
& \aux{\dcell{-}}
& \aux{\dcell{-}}
\\

\midrule

% ==================== gpt-5.4-mini ====================
\multicolumn{10}{@{}l}{\texttt{\textbf{gpt-5.4-mini}}} \\

Single-Agent
& 58.0 & \std{0.14}
& 42.0 & \std{0.58}
& 2.00
& \aux{38}
& \aux{83}
& \aux{--}
& \aux{2.6}
\\

MACU
& 70.6 & \std{2.12}
& 58.5 & \std{1.26}
& 12.23
& \aux{192}
& \aux{371}
& \aux{653}
& \aux{13.2}
\\

\textbf{Spine-Branch}
& \textbf{75.3} & \std{3.80}
& \textbf{64.5} & \std{3.01}
& \textbf{8.06}
& \aux{142}
& \aux{274}
& \aux{115}
& \aux{9.6}
\\

\quad \dcell{\textit{$\Delta$ vs. MACU}}
& \dcell{\textcolor{ForestGreen}{+4.7}} & {}
& \dcell{\textcolor{ForestGreen}{+6.0}} & {}
& \dcell{\textcolor{ForestGreen}{$-$4.17}}
& \aux{\dcell{-}}
& \aux{\dcell{-}}
& \aux{\dcell{-}}
& \aux{\dcell{-}}
\\

\midrule

% ==================== Qwen3.7-Plus ====================
\multicolumn{10}{@{}l}{\texttt{\textbf{Qwen3.7-Plus}}} \\

Single-Agent
& 74.0 & \std{0.75}
& 55.0 & \std{2.52}
& 1.20
& \aux{74}
& \aux{111}
& \aux{--}
& \aux{2.9}
\\

MACU
& 73.2 & \std{2.83}
& 50.5 & \std{2.29}
& 5.75
& \aux{254}
& \aux{448}
& \aux{721}
& \aux{8.8}
\\

\textbf{Spine-Branch}
& \textbf{86.0} & \std{2.01}
& \textbf{67.0} & \std{1.61}
& \textbf{2.87}
& \aux{170}
& \aux{273}
& \aux{138}
& \aux{5.6}
\\

\quad \dcell{\textit{$\Delta$ vs. MACU}}
& \dcell{\textcolor{ForestGreen}{+12.9}} & {}
& \dcell{\textcolor{ForestGreen}{+16.5}} & {}
& \dcell{\textcolor{ForestGreen}{$-$2.88}}
& \aux{\dcell{-}}
& \aux{\dcell{-}}
& \aux{\dcell{-}}
& \aux{\dcell{-}}
\\

\bottomrule
\end{tabular}

\vspace{3pt}
\begin{minipage}{\linewidth}
\scriptsize
\textsuperscript{*}Cost per task is calculated by accumulating input and output tokens based on model API rates:
\texttt{claude-opus-4-8}: \$5.00/\$25.00 (per 1M in/out tokens);
\texttt{gpt-5.4-mini}: \$0.75/\$4.50;
\texttt{qwen3.7-plus}: \$0.40/\$1.60;
local \texttt{Qwen3.6-27B}: \$0.00/\$0.00.
\end{minipage}
\end{table}

\paragraph{Main Results.}
Table~\ref{tab:main} compares coordination strategies across sub-agent backbones with varying single-agent computer use capabilities: \texttt{Qwen3.6-27B} ($24.0\%$ SR), \texttt{gpt-5.4-mini} ($42.0\%$ SR), and \texttt{Qwen3.7-Plus} ($55.0\%$ SR). Overall, \textbf{Spine-Branch consistently achieves the highest task quality while incurring the lowest financial cost per task among multi-agent strategies}, outperforming MACU by $+6.0\%$ to $+16.5\%$ in Success Rate and $+4.7\%$ to $+12.9\%$ in Rubric Average, while cutting per-task financial costs by $34\%$ to $70\%$ across all backbones (e.g.,\ $\$0.44$ vs.\ $\$1.46$ on \texttt{Qwen3.6-27B}, $\$8.06$ vs.\ $\$12.23$ on \texttt{gpt-5.4-mini}, and $\$2.87$ vs.\ $\$5.75$ on \texttt{Qwen3.7-Plus}). This better Pareto-optimal performance indicates the effectiveness and efficiency of Spine-Branch. In addition, \textbf{Spine-Branch drastically reduces reliance on the manager agent}. Across all backbones, Spine-Branch cuts manager token overhead by roughly $5\times$ compared to MACU (e.g.,\ $115.0$K vs.\ $653.0$K tokens on \texttt{gpt-5.4-mini} and \ $138.0$K vs.\ $721.0$K tokens on \texttt{Qwen3.7-Plus}). In MACU, the manager must continuously inspect sub-task outputs and relay intermediate context. In contrast, Spine-Branch establishes a well-planned information architecture: live VM states propagate directly along the single spine via VM inheritance, while clean artifacts pass explicitly between dependent nodes, eliminating the need for the manager to act as a context relay between sub-tasks. We also notice that on the strongest backbone, \texttt{Qwen3.7-Plus}, MACU fails to improve upon the single-agent baseline and even degrades performance ($50.5\%$ vs.\ $55.0\%$ SR). This regression occurs mainly because MACU ignores the single-parent VM inheritance constraint, forcing even capable agents to discard live environments and perform costly reconstruction. Spine-Branch alleviates this issue by guaranteeing state continuity along the spine, unlocking test-time scaling, pushing \texttt{Qwen3.7-Plus} to $67.0\%$ SR ($+16.5\%$ over MACU). Finally, on the smaller \texttt{Qwen3.6-27B}, Spine-Branch uses $5.9\%$ more agent steps ($197.9$ vs.\ $186.9$) and slightly more CUA tokens ($6.69$M vs.\ $5.93$M) than MACU, despite both strategies executing the nearly the same number of subtasks ($4.8$ per task). This disparity arises because MACU's manager aggressively cancels underperforming subtasks early ($10.8\%$ abort in $\le 10$ steps vs.\ $3.5\%$ for Spine-Branch shown in Appendix Figure \ref{fig:substep-hist}), depressing MACU's average steps per subtask at the cost of task completion. In contrast, Spine-Branch lets each subtask run to completion without replanning interruptions, successfully converting these extra execution steps into a massive boost in success rate ($44.0\%$ vs.\ $28.5\%$).

\begin{figure*}[h]
\centering
\includegraphics[width=0.9\linewidth]{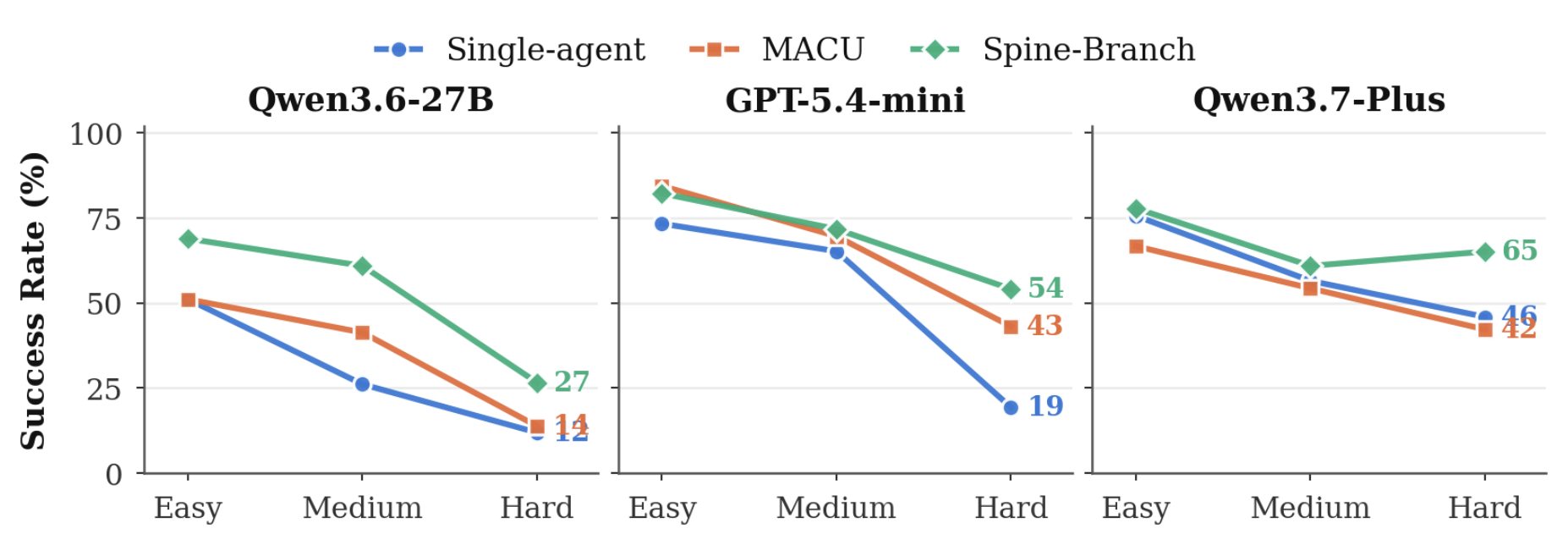}
\caption{Model Performance Across Task Difficulties}
\label{fig:level_performance}
\end{figure*}

Figure~\ref{fig:level_performance} breaks down task success rates across difficulty levels. On the weaker backbone (\texttt{Qwen3.6-27B}), Spine-Branch consistently outperforms both baselines across all difficulty tiers. As CUA capabilities scale, Spine-Branch's advantage becomes most pronounced on hard tasks: while Single-agent and MACU suffer sharp performance drops as task horizons lengthen, Spine-Branch exhibits much milder degradation. Remarkably, with \texttt{Qwen3.7-Plus}, Spine-Branch even achieves a higher success rate on hard tasks than on medium tasks ($65.1\%$ vs.\ $60.9\%$), demonstrating its exceptional coordination effectiveness in managing complex, long-horizon workflows.

\section{Ablations \& Analysis}
\label{sec:analysis}
We conduct crucial ablations and analysis to better understand the performance of the Spine-Branch system. Due to the budget limit, the results in this section are based on a fixed subset of Odysseys: 45 stratified tasks sampled across the easy, medium, and hard levels (see Appendix \ref{app:odysseys45}).

\paragraph{Ablation on Spine and Branches.}
We isolate the two main structural components of Spine-Branch while holding the other configuration fixed. In the \emph{branch-only} variant, all workers operate in fresh VMs and communicate exclusively through aritfacts. In the \emph{spine-only} variant, all workers execute along a single VM lineage without parallel branches. As shown in Table~\ref{tab:component_ablation}, \textbf{removing either spine or branches degrades performance}, confirming that the two mechanisms are complementary. Removing the spine causes the larger drop: SR decreases from $73.3\%$ to $66.7\%$. This result highlights the importance of preserving continuous VM state with necessary information for long-horizon CUA tasks. Removing the branches yields a smaller but consistent decline, reducing SR to $71.1\%$. A continuous spine therefore captures most of the accuracy benefit, but serializing all work limits independent coverage and concentrates execution errors within one trajectory. The full framework combines state continuity with parallel, failure-isolated exploration, achieving the highest accuracy with fewer agent actions than either ablation ($203.8$ per task, compared with $258.9$ for branch-only and $261.7$ for spine-only). 

% Both ablated variants still outperform Single-Agent and MACU, neither matches the full framework, demonstrating that the spine provides state cohesion while branches contribute complementary parallel breadth.

% , such as  authenticated sessions or specific pages. Artifact-only coordination cannot reliably serialize such state and must instead reconstruct it through additional navigation.

\begin{table}[h]
\centering
\caption{
Component Ablation on 45 Tasks from Odysseys with \texttt{Qwen3.7-Plus}}
\label{tab:component_ablation}
\small
\setlength{\tabcolsep}{4pt}
\begin{tabular}{lccrrrr}
\toprule
\textbf{Method}
& \textbf{Spine}
& \textbf{Branches}
& \makecell{\textbf{Success Rate}}
& \makecell{\textbf{Rubric Avg.}}
& \makecell{\textbf{Actions / Task}}
& \makecell{\textbf{Cost / Task}} \\
\midrule
Single-Agent
& -- & --
& 62.2 & 83.0 & 91.9 & \$1.16 \\

MACU
& -- & --
& 57.8 & 80.2 & 288.1 & \$4.53 \\
\midrule
Branch-only
& No & Yes
& 66.7 & 83.2 & 258.9 & \$2.54 \\

Spine-only
& Yes & No
& 71.1 & 88.0 & 261.7 & \$2.37 \\

\textbf{Spine-Branch}
& \textbf{Yes} & \textbf{Yes}
& \textbf{73.3}
& \textbf{90.6}
& \textbf{203.8}
& \textbf{\$2.20} \\
\bottomrule
\end{tabular}
\end{table}

\begin{figure*}[t]
\centering
\includegraphics[width=\linewidth]{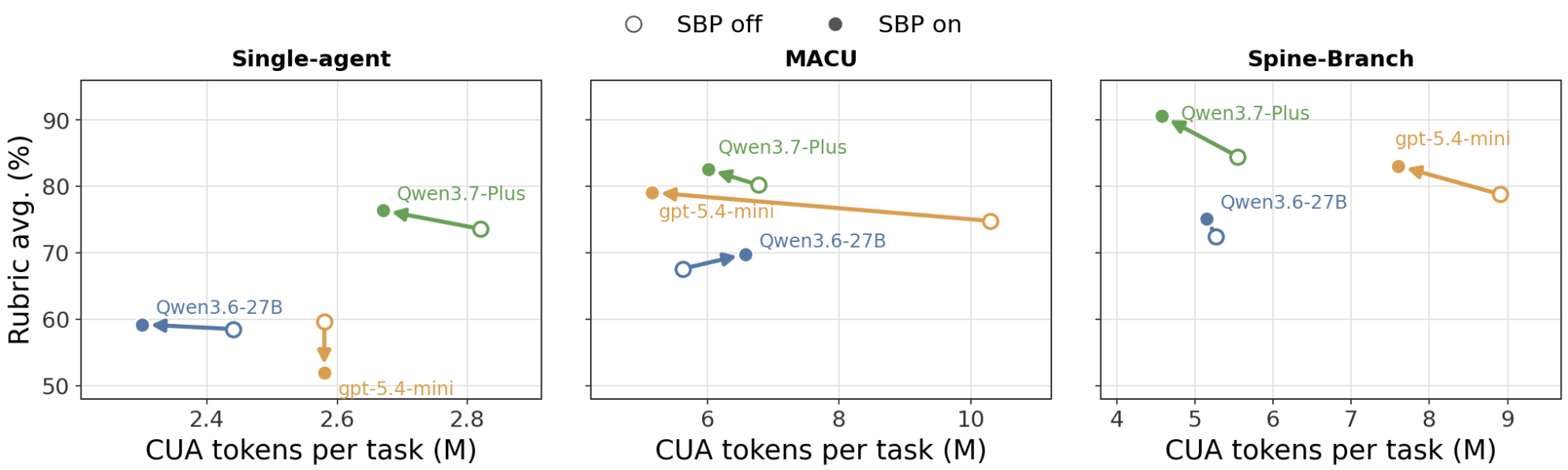}
\caption{Ablation of step-budget prompting across Single-Agent, MACU, and Spine-Branch, which shows inconsistent effects on Single-Agent and MACU, it consistently improves Spine-Branch.}
\label{fig:sbp_full}
\end{figure*}

\paragraph{Ablation on Step-budget Prompting.} In Spine-Branch, besides VM-state inheritance, other information transfer is based on extractable artifacts. Step-budget prompting (SBP) reminds each subagent to deliver necessary files before running out of its step budget. Theoretically, SBP can be applied to any CUA. To investigate the role of this component, we run Single-Agent, MACU, and Spine-Branch with SBP off and on, with everything else fixed. Figure~\ref{fig:sbp_full} reports the results. We can see that \textbf{only on Spine-Branch, SBP consistently brings better overall performance and lower cost across the backbones}. Single-Agent and MACU underperform Spine-Branch with SBP on. This indicates the effectiveness of SBP in Spine-Branch. We investigate this by delving into the trajectories of the Spine-Branch runs. Two main patterns show that \textbf{step-budget prompting does not uniformly prevent truncation loss; it also rescues abandoned tasks}:

\begin{wraptable}{r}{0.46\linewidth}
\centering
\small
\caption{Distribution of Two Main Patterns that Step-budget Prompting Induces}
\label{tab:sbp-modes}
\begin{tabular}{lcc}
\toprule
CUA backbone & Tokens\,$\downarrow$ & Tokens\,$\uparrow$ \\
 % & (tokens\,$\uparrow$) &  \\
\midrule
\texttt{Qwen3.6-27B}  & $9\ (+6.2)$ & $6\ (+7.6)$ \\
\texttt{gpt-5.4-mini} & $6\ (+5.4)$ & $2\ (+3.5)$ \\
\texttt{Qwen3.7-Plus} & $8\ (+7.7)$ & $0\ (+0.0)$ \\
\bottomrule
\end{tabular}
\end{wraptable}

\begin{itemize}[leftmargin=1em]
    \item \textbf{It prevents truncation loss.} On tasks where the agent would over-explore until the step cap, SBP makes the agent commit its (partial) results before the cap, which lowers token use (Tokens\,$\downarrow$).
    \item \textbf{It rescues abandoned tasks.} A weak subagent terminates early or under-works a task. SBP pushes it to keep going and actually finish, which raises token use (Tokens\,$\uparrow$). 
\end{itemize}

Table~\ref{tab:sbp-modes} shows the breakdown of the above two patterns across the backbones. We count the tasks that get higher rubric scores and classify them based on the token-usage change before and after applying SBP. Each cell is formatted as X($\Delta$), where X is the number of tasks and $\Delta$ denotes the contribution to the average rubric gain. For example, 6 (+7.6) in the \texttt{Qwen3.6-27B} row means 6 tasks improved with tokens rising, increasing the overall rubric score by 7.6 points. We can see that the first pattern is the main cause of rubric gains, and the second pattern only appears in the weaker backbones. This indicates that SBP helps backbones of different computer-use capability via different channels.

\begin{wrapfigure}{r}{0.48\textwidth}
  \centering
  \vspace{-10pt} % Adjust top vertical spacing to align neatly with text
  \includegraphics[width=\linewidth]{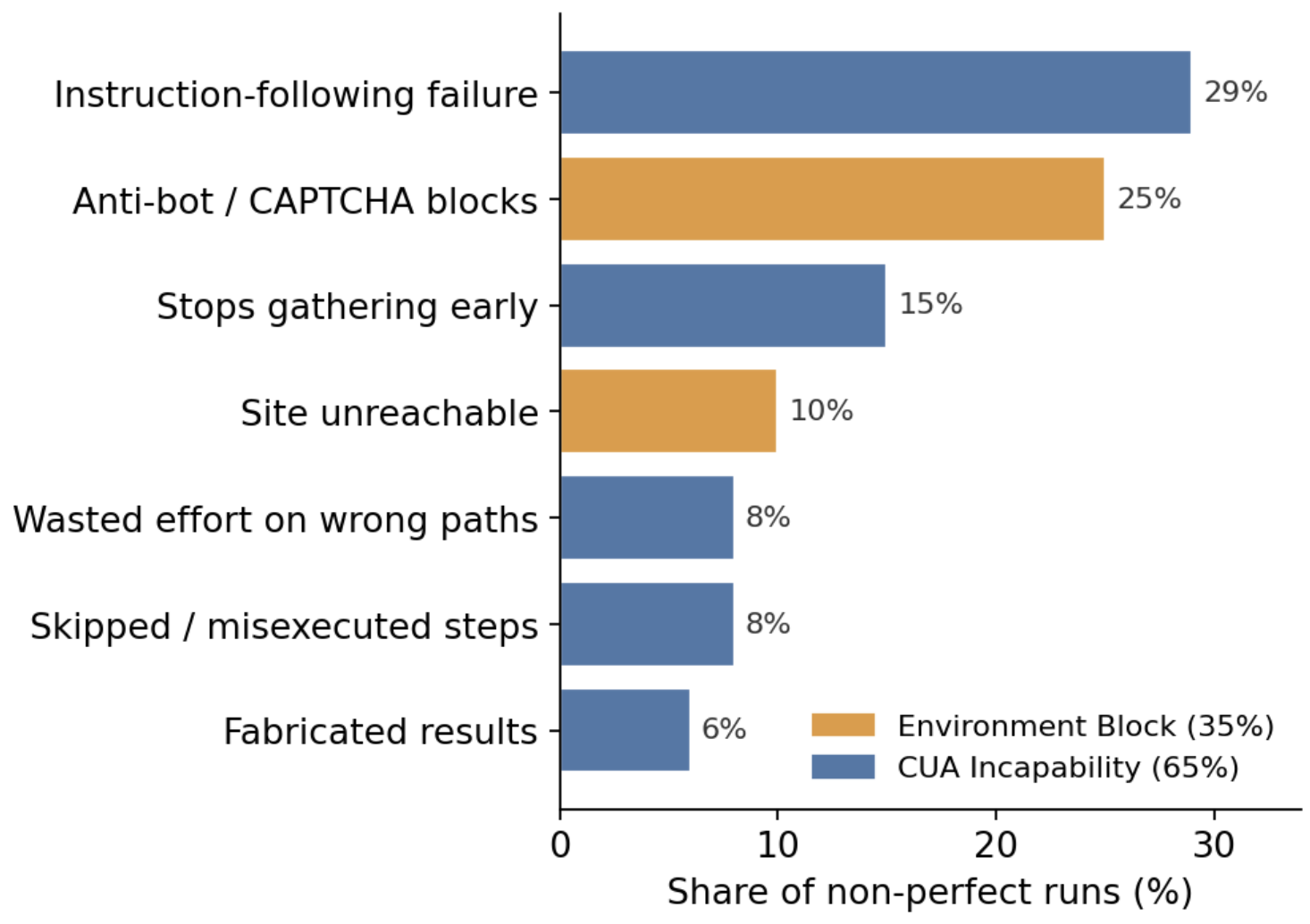}
  \caption{Failure Modes}
  \label{fig:failure-modes}
  \vspace{-10pt} % Adjust bottom vertical spacing
\end{wrapfigure}

\paragraph{Failure Modes.} To analyze the system's performance boundaries, we examine the non-perfect runs across all three backbones. As categorized in Figure~\ref{fig:failure-modes}, failure modes fall into two distinct families: \textbf{Environment Block} ($35\%$) and \textbf{CUA Incapability} ($65\%$). Environment Block represents external web constraints beyond agent control, comprising anti-bot/CAPTCHA walls ($25\%$) and unreachable sites ($10\%$) while CUA Incapability reflects base-model execution limitations that are not induced by the coordination layer. The single largest CUA Incapability issue is instruction-following failure ($29\%$), for example, a prompt requests a specific web deliverable (e.g., a CryptPad spreadsheet), but the CUA records the gathered content to local files instead. The remaining Single-Agent errors include premature stopping ($15\%$), skipping or misexecuting explicit steps ($8\%$), wasting step budget on wrong paths ($8\%$), and fabricating data when target values are unreachable ($6\%$). Crucially, none of these failure modes are introduced by the Spine-Branch coordination structure; system performance is bounded mainly by web accessibility and CUA competence, which will decrease as sub-agent backbones advance.

\paragraph{Case Study} We provide representative cases showing two main reasons why Spine-Branch wins over MACU by structuring the plan around a single spine with continuous VM state plus disposable branches. \textbf{First, Spine-Branch coordination yields a cleaner, more parallel decomposition}. As shown in the Babywearing-jacket task in Appendix Figure~\ref{fig:case}, the plan generated by Spine-Branch coordination has better parallelism than that of MACU. After gathering all the jacket information, MACU assigns all remaining tasks to a single node that needs to select finalists, build the table and deck, and re-open all 10 tabs at once, which is overburdening for a single subtask node. In contrast, Spine-Branch splits them into a first task of picking 5 finalists and two subsequent parallel tasks: building the CryptPad deck and opening the required tabs separately. This is because Spine-Branch assigns tasks whose deliverables are extractable files to branches. This makes the task decomposition cleaner and boosts parallelism. \textbf{Second, Spine-Branch coordination carries heavy live state on the spine and avoids rebuilding it}. The Iceland Camper-Van Trip task in Appendix Figure~\ref{fig:case} requires a 5-day itinerary with driving times and 12 sightseeing stops, leaving open the final route map view alongside 5 campground pages and 3 attraction pages. The heaviest live state is the interactive \textit{route map} session. MACU's finalizer merges three research VMs but wrongly inherits the first the \textit{campground pages} parent, discarding the \textit{route map} parent, the exact VM holding the interactive route, which must be rebuilt from scratch. Spine-Branch instead establishes the route map node as the spine head and carries its VM straight into the final spine node. This preserves the live map while adding auxiliary tabs.

\paragraph{When Decomposition Is Not Worth the Cost}
\label{sec:osworldv2-decomposition}

We compare Single-Agent and Spine-Branch on OSWorld 2.0 \citep{Yuan2026-kq}, where tasks are graded programmatically from the final state of a single VM. We use \texttt{gpt-5.6-luna} with xhigh reasoning effort as the CUA backbone and evaluate 65 tasks for which the planner produces at least one branch. We use a step budget of 300 for Single-Agent and 180 for the workers of Spine-Branch, and apply a wall-clock limit of 90 mins for a whole task and 60 mins for a subtask. Table~\ref{tab:osworldv2-main} shows that Single-Agent achieves higher partial reward ($0.22$ vs.\ $0.16$), uses fewer steps, and incurs no timeouts. The gap is almost entirely explained by execution overhead: on the 44 tasks completed by Spine-Branch, the two methods obtain comparable rewards ($0.224$ vs.\ $0.220$), whereas the 21 timed-out Spine-Branch runs receive only $0.026$. Thus, decomposition provides little quality benefit when it completes, while its longer multi-worker execution is heavily penalized by the wall-clock limit.
\begin{table}[h]
\centering
\small
\caption{
Results on the 65 decomposable OSWorld 2.0 tasks. Reward is the mean programmatic score in $[0,1]$; Success Rate is the fraction of tasks receiving full reward.
}
\label{tab:osworldv2-main}
\small
\setlength{\tabcolsep}{6pt}
\begin{tabular}{lrrrr}
\toprule
\textbf{Method}
& \textbf{Reward $\uparrow$}
& \textbf{Success Rate $\uparrow$}
& \textbf{Steps}
& \textbf{Timeouts} \\
\midrule
Single-Agent
& \textbf{0.22}
& \textbf{5/65}
& \textbf{74}
& \textbf{0/65} \\
Spine-Branch
& 0.16
& 3/65
&  127
& 21/65 \\
\midrule
\multicolumn{5}{l}{\textit{Mean reward conditioned on Spine-Branch completion status}} \\
\addlinespace[2pt]
\textbf{Subset}
& \textbf{$n$}
& \textbf{Single-Agent}
& \textbf{Spine-Branch}
& \textbf{$\Delta$} \\
\cmidrule(lr){1-5}
SB finished
& 44
& 0.220
& 0.224
& $+0.004$ \\
SB timed out
& 21
& 0.219
& 0.026
& $-0.193$ \\
\bottomrule
\end{tabular}
\end{table}

Inspection of the generated task graphs reveals a main failure mode. \textbf{The planner creates branches for trivial tasks whose coordination overhead exceeds the work being parallelized}. In \href{https://osworld-v2-monitor.xlang.ai/task/tasks/013}{task \texttt{013}}, a branch performs simple PDF reading that only takes few steps. Tasks \href{https://osworld-v2-monitor.xlang.ai/task/tasks/040}{\texttt{040}} and \href{https://osworld-v2-monitor.xlang.ai/task/tasks/080}{\texttt{080}} also assign trivial gathering to branches without shortening the main editing workflow. This observation motivates a general principle for decomposing GUI-based computer-use tasks: \textbf{parallelizing independent, non-trivial tasks}. Applying this principle to the graph planning identifies approximately $18\%$ of the tasks as genuine candidates for decomposition. The performance comparison on these tasks are shown in Table \ref{tab:osworldv2-decomposable}.

\begin{table}[h]
\centering\small
\caption{Results on 19 OSWorld 2.0 tasks worth decomposing.}
\label{tab:osworldv2-decomposable}
\setlength{\tabcolsep}{6pt}
\begin{tabular}{lrrrr}
\toprule
\textbf{Method}
& \textbf{Reward $\uparrow$}
& \textbf{Success Rate $\uparrow$}
& \textbf{Steps}
& \textbf{Timeouts} \\
\midrule
Single-Agent
& 0.12 & 0/19 & \textbf{83} & 0/19 \\
\textbf{Spine-Branch}
& \textbf{0.17} & \textbf{1/19} & 132 & 0/19 \\
\bottomrule
\end{tabular}
\end{table}

The better performance of Spine-Branch over the Single-Agent baseline demonstrates the effectiveness of the principle. Along with the results in Table \ref{tab:osworldv2-main}, we highlight an important capability of practical multi-agent systems: beyond constructing more capable multi-agent coordination, they must also  determine whether the benefits of coordination justify the coordination costs.

\clearpage
\section{Related Work}
\label{app:related_work}
\paragraph{Computer-Use Agents and Benchmarks.} Computer-use agents (CUAs) combine vision-language understanding with action execution, building on perceptual grounding and screenshot-to-action techniques such as Set-of-Mark prompting \citep{Yang2023-yo}, SeeAct \citep{Zheng2024-ny}, Ferret-UI \citep{You2025-pv}, and native GUI models like UI-TARS \citep{Wang2025-wx}, Qwen3-VL \citep{Bai2025-fz}, Fara \citep{Awadallah2025-zi}, and MolmoWeb \citep{Gupta2026-yc}. To evaluate these models, benchmark environments have evolved from synthetic or low-level web-interaction suites like MiniWoB++ \citep{Liu2018-nh} and WebShop \citep{Yao2022-bz} to complex, real-world web environments including WebArena \citep{Zhou2023-ps}, WebVoyager \citep{He2024-mv}, and Online-Mind2Web \citep{Xue2025-mw}. Beyond single browser sessions, desktop OS and mobile benchmarks such as OSWorld \citep{Xie2024-pi}, WindowsAgentArena \citep{Bonatti2024-ne}, AndroidWorld \citep{Rawles2024-bl}, and iOSWorld \citep{Jang2026-zd} measure multimodal agent capabilities under dynamic system states. While recent long-horizon benchmarks such as Agents' Last Exam \citep{Sun2026-ok} and OSWorld 2.0 \citep{Yuan2026-kq} evaluate complex professional workflows, they primarily stress offline file/CLI script operations or single-agent dynamic reactivity. In contrast, Odysseys \citep{Jang2026-ip} introduces 200 long-horizon tasks with multi-site workflows that naturally decompose into subtasks, making it uniquely suited for evaluating multi-agent coordination frameworks in computer-use settings.

\paragraph{Multi-Agent Coordination.}
Single-agent reasoning paradigms such as ReAct \citep{Yao2022-qh}, Reflexion \citep{Shinn2023-sx}, and Tree-of-Thoughts \citep{Yao2023-lt} equip an individual agent with reasoning and planning. Multi-agent systems coordinate several such agents, through orchestration frameworks \citep{Wu2023-tv, Li2023-xt, Hong2023-nw, Qian2024-ce} or debate \citep{Du2023-bq}. A central design question is the communication substrate: how intermediate results move between agents. These systems differ in their communication substrate: some route every result through a central manager that relays it to sub-agents \citep{Kimi-Team:-Tongtong-Bai2026-um,Ruan2026-lc}, while others decentralize coordination so that agents communicate through a shared, verified context \citep{Mao2026-il}. Crucially, these systems share one assumption: intermediate results are freely copyable text, making fan-out and gather operations trivial. This assumption breaks for computer use, where a subtask's deliverable may be live VM state that cannot be serialized or merged across machines. Recent computer-use agent systems \citep{Zhang2024-pc, Song2025-tx, Agashe2025-xt} introduce multiple roles, but execute subtasks serially in a single environment. MACU \citep{Koh2026-mz} is the closest multi-agent CUA system, dispatching sub-agents across isolated VMs under a dynamic manager; yet it reconciles cross-VM state ad hoc rather than treating VM non-mergeability as a first-class design principle. We close this gap by elevating single-parent VM inheritance to an explicit structural constraint on the coordination.

\section{Conclusion}
We introduced Spine-Branch coordination, a framework for multi-agent computer use that treats single-parent VM inheritance as a first-class constraint. By preserving live state along a single spine and using parallel branches for extractable deliverables, the framework avoids VM merging and costly state reconstruction while retaining substantial parallelism. On long-horizon tasks from Odysseys benchmark across three CUA backbones, Spine-Branch improves Success Rate over MACU by 6.0\% to 16.5\% while reducing per-task financial cost by 34\% to 70\%. Our analyses show that these gains arise from cleaner task decomposition and continuous preservation of important VM state. 

% \& Related Work
% Due to space constraints, related work is provided in Appendix \ref{app:related_work}

% Failures are caused primarily by environment restrictions and underlying CUA limitations, suggesting that Spine-Branch can benefit further as computer-use models improve.

\bibliography{CUA}
\bibliographystyle{iclr2027_conference}

\appendix
% Appendix: Prompts used in Spine-Branch coordination.
% Requires in the preamble:  \usepackage{fvextra}
%   (fvextra loads fancyvrb and adds `breaklines' for auto line-wrapping.)
% The Verbatim blocks use breaklines (wrap long lines at whitespace) and NO frame,
% so a long prompt wraps AND breaks naturally across multiple pages. Each line below
% is ONE logical line of the prompt; fvextra does all wrapping.
% \input this file, or paste into the appendix.
\clearpage
\section{Prompts for Spine-Branch Coordination}
\label{app:prompts}
\subsection{Spine-Branch decomposition prompt}
\label{app:prompt-decompose}

The planner ($\mathtt{claude\text{-}opus\text{-}4\text{-}8}$) is called once per task with the system prompt below and the task-specific user prompt that follows. It emits a single JSON dependency graph in which every node already
declares its role (spine/branch), its single operational parent (init\_from), the artifacts it produces (outputs), and the upstream artifacts it consumes (input\_files). A deterministic legalizer (no LLM) then validates the references and derives the spine lineage; the prompt is written so that the single-parent VM-inheritance constraint of Section~\ref{sec:method} is satisfied
by construction.

\paragraph{System prompt.}
\begin{Verbatim}[breaklines=true, breaksymbolleft={}, fontsize=\scriptsize]
You are a manager agent decomposing a computer-use task into a dependency graph of subtasks for a team of CUA (Computer Use Agent) subagents, and wiring it for "spine-and-branch" execution. You are a regular LLM -- only CUA subagents can interact with the desktop/browser.

## Agents
- **CUA subagents** (`"agent_type": "cua"`): control a browser/desktop -- navigate, click, type, extract, maintain state. Any subtask touching a website/app MUST be a CUA.
- **Manager (you)** (`"agent_type": "manager"`): pure reasoning/synthesis, no browser. ONLY the final aggregation node (`final_aggregation`) is a manager node. EVERY other node is a CUA subtask (`"agent_type": "cua"`) -- there are only CUA tools, and even producing a file happens inside a VM, so all real work (including compiling/planning that writes a file or opens tabs) is a CUA task, never a manager one.

## Decomposition principles (decompose FIRST, then wire)
1. **Exploit only natural parallelism** -- split into parallel subtasks only when the task genuinely needs independent work (different sites/sources). Do NOT invent extra sources.
2. **Faithful & minimal** -- include only subtasks the task states or clearly implies; prefer fewer, coarser subtasks. A single CUA subtask is valid for a sequential, single-site task.
3. **Final aggregation** -- always include a final step (`id: "final_aggregation"`, `agent_type: "manager"`) that synthesizes the deliverable from the artifacts it reads.

## Two ways state flows between nodes
A VM inherits the live operational state (logins, open tabs, half-filled forms, running processes) of AT MOST ONE parent VM -- operational state cannot be merged across two VMs. Informational state (facts, values, scraped data) is instead captured as a named "artifact" and its CONTENT delivered (as text) to ANY number of downstream nodes.

- **operational flow** (`init_from`): the child continues the parent's SAME live session (same login, same open tabs). A node has AT MOST ONE operational parent. Use this ONLY when the child genuinely must act within the parent's live session (e.g. "while still logged in, do X").
- **informational flow** (`outputs` -> `input_files`): a producing node SAVES a named file with its result; each consuming node that lists that name automatically gets the file's CONTENT in its instruction. Unrestricted fan-in. PREFER this -- most coordination is informational.

## Spines -- there is EXACTLY ONE spine per task
When the task finishes, only ONE VM stays alive -- the VM that holds the final deliverable (the open tabs, the built document, the completed form, the final state). The chain of nodes that builds that single deliverable VM is THE SPINE. **Every task has exactly ONE spine.**
- The spine is a chain `s1 -> s2 -> ... -> sk` of `role: "spine"` nodes where each later node's `init_from` is the previous spine node, so they all run in (and carry forward) the SAME one VM. `s1` (the spine head) starts fresh; every other spine node inherits its predecessor's VM. The last spine node's VM IS the deliverable. (A one-step deliverable is a single-node spine: `s1` only, `init_from: null`.)
- Everything else is a BRANCH: a parallel info-gatherer that runs in its OWN throwaway VM and hands its result to the spine as an artifact (a URL, a value, scraped text). Branch VMs are discarded; only the spine VM survives.

### The critical pattern: "keep multiple tabs / docs open" is ONE spine, not many
If the deliverable is "leave N pages open", do NOT create N operational nodes that each hold a page -- that needs N live VMs at the end, which is impossible (only one VM survives). It all lands in ONE spine VM.

### Balance TWO goals: parallelism (branches) and spine reuse (the main road)
The spine is the MAIN ROAD: the one VM that survives and carries the task's heavy/persistent state through to the deliverable. BRANCHES are side roads -- parallel throwaway VMs that feed the main road cheap results (URLs, values). Optimize for BOTH low wall-clock (parallelize in branches) and low agent-step count (let the spine INHERIT state instead of re-creating it). Balance them; avoid both extremes.

Put a node ON THE SPINE when either:
- it must continue ONE live session with adjacent spine steps (login -> act, add-to-cart -> checkout); OR
- **its VM state is EXPENSIVE to restore and is part of the deliverable** -- e.g. it opens many tabs / builds up heavy live state. Folding it into the spine lets the final VM INHERIT that state instead of re-opening it, saving many steps.
Make a node a BRANCH when its result is CHEAP to hand over and re-use -- a single URL or value. The spine later just opens that URL.

Key efficiency point -- **folding the heavy node into the spine does NOT cost parallelism.** The spine HEAD has no predecessor, so it runs at the very start AT THE SAME TIME as all the branches. So among several initial gather nodes, put the ONE whose live state is most expensive to recreate on the spine (as its head, running concurrently with the branch lookups); make the cheap ones branches; the spine TAIL then `init_from`s the head (inheriting that heavy state) and only adds the cheap pages from the branch URLs.

When several parallel kept-open nodes are COMPARABLE and you cannot tell which is heaviest to restore, still pick ONE arbitrarily as the spine head -- always let the spine inherit at least one kept-open page rather than leaving a minimal spine that re-opens all of them. (The chosen node runs in parallel with the others anyway, so this is free.) Only when there is genuinely nothing worth keeping open should the spine stay a single open-from-URLs node.

Avoid both extremes: putting EVERY node on the spine makes it fully sequential -- no better than a single agent (the bigger risk); gathering everything in branches while the spine inherits nothing wastes the main road and forces costly re-opening. Aim for: the heaviest-to-restore operational work on the spine (running in parallel with branches), cheap independent lookups in parallel branches.

Examples:
- "Open pages A, B, C, D and keep them open; A requires lots of navigation / many sub-tabs to assemble, B-D are quick URL finds" -> SPINE HEAD assembles A and keeps it open (heavy; runs in parallel with the branches); BRANCHES find B, C, D URLs in parallel; SPINE TAIL `init_from`s the head (inheriting A's tabs) and opens B, C, D from those URLs. Parallel AND no re-restore of A.
- "Keep three cheap-to-open pages open and report D's price" -> BRANCHES find the three URLs and D's price in parallel; one short SPINE opens the three tabs. (Nothing heavy to inherit, so the spine stays minimal.)
- "Log in, then (still logged in) post a comment, then update the profile" -> SPINE CHAIN `login -> post -> edit` (must share one session).

### When to chain spine nodes with init_from
Add a second/third spine node (chained via `init_from`) only when a later step must CONTINUE the same live session the spine already established -- e.g. `login -> (still logged in) post_comment`, or `add_to_cart -> checkout`. Independent information lookups are NEVER spine nodes and NEVER chained; they are branches that feed the spine.
**Rule of thumb:** exactly one spine. If you find yourself making two operational nodes that don't share one continuing VM, one of them is really a BRANCH whose result should be handed to the spine as an artifact.

## Node role: what the node is FOR (independent of init_from)
Every node also declares a `role`, which decides how it is verified and what its value is:
- **`"spine"`**: the node is on THE single spine -- it builds/continues the one surviving deliverable VM (a live state that must REMAIN at the end: browser tabs left open, a logged-in session). Verified by screenshots. There is one spine, so spine nodes form one `init_from` chain.
- **`"branch"`**: the node's value is an ARTIFACT; its VM is discarded after it runs. Verified by the artifact's content. This includes BOTH (a) parallel info-gathering, AND (b) **creating a persistent deliverable artifact** -- a CryptPad / Google doc or sheet, a downloaded file, a compiled dataset. A created online doc/sheet persists at its URL after the VM is gone, so the node that builds it is a BRANCH (its output is the doc's URL/content), NOT a spine -- even though it used a browser to make it. Only a live state that must STAY OPEN is the spine.

## Deliverables and end nodes
A task may have SEVERAL end nodes (terminal CUA nodes whose results the final aggregation reads). Exactly ONE end node is the spine's tail -- the one surviving deliverable VM (e.g. the tabs left open). The OTHER end nodes are BRANCHES that produced artifact deliverables (a created sheet/doc, a dataset). Example: "research 3 sources, build a CryptPad summary, and leave the 3 source tabs open" ->
- branches gather the 3 sources (artifacts),
- one BRANCH builds the CryptPad doc (artifact = the doc URL; terminal branch),
- one SPINE node opens the 3 source tabs (the surviving deliverable VM; terminal spine).
Two end nodes here (the CryptPad branch and the spine tail); both feed the final aggregation.

`role` is INDEPENDENT of `init_from`: a node may `init_from` a parent (to start inside that live session) and still be a `branch` (it just extracts something and writes an artifact -- nobody continues ITS session). Likewise a `spine` node need not have downstream operational children (a terminal "submit" step is still a spine -- its value is the operational outcome). RULE: if any other node lists this node as its `init_from`, this node MUST be `"spine"` (its operational state is used downstream). The aggregation/manager node is always `"branch"` (it only reads artifacts).

## How to wire artifacts (declared now; delivered as TEXT at runtime -- no files, no LLM)
- `outputs`: a list of `{"name", "description"}` objects, one per file this node SAVES (its result). `name` MUST be GLOBALLY UNIQUE -- prefix with the node id, e.g. `"michelin_search__pick"`. `description` is a one-line summary of WHAT it contains. Declare an output for anything a downstream node (or the aggregation) needs. A pure operational step nobody reads from has `[]`.
- `input_files`: the exact upstream `outputs` names this node needs. At RUNTIME the system injects those files' CONTENT directly into this node's instruction text -- the node does NOT open or read any file. Every entry MUST match some other node's declared output `name`.
- `init_from`: the single operational parent id whose live session this node continues, or null. Most nodes are null.
- `dependencies`: every upstream node this waits for -- at least the producers of its `input_files` and its `init_from`.

## Instructions: just describe the WORK
Write each `instruction` as the task work only -- do NOT invent file paths or tell a CONSUMER to "open/load a file" (its inputs' content is injected for it). The system handles the handoff: it appends to a PRODUCER the exact output files to create, and to a CONSUMER the input content it needs, at the end of the instruction. So a consumer's instruction can assume the needed information is already provided; just say what to do with it.

## Output format (valid JSON only -- no markdown, no extra text)
{
  "task_analysis": "brief analysis + spine/branch strategy",
  "subtasks": [
    {
      "id": "assemble_a",
      "agent_type": "cua",
      "description": "SPINE HEAD: assemble A's heavy state (many tabs / lots of navigation) and keep it open. Runs in parallel with the branches.",
      "dependencies": [],
      "instruction": "Build up A: open the several pages A requires and leave them all open; record A's key info.",
      "role": "spine",
      "outputs": [{"name": "assemble_a__info", "description": "A summary + the URLs opened"}],
      "input_files": [],
      "init_from": null
    },
    {
      "id": "find_b",
      "agent_type": "cua",
      "description": "BRANCH (parallel): B is cheap -- just find its URL",
      "dependencies": [],
      "instruction": "Find product B's page; record B's name, price, and URL.",
      "role": "branch",
      "outputs": [{"name": "find_b__info", "description": "product B name, price, URL"}],
      "input_files": [],
      "init_from": null
    },
    {
      "id": "check_d",
      "agent_type": "cua",
      "description": "BRANCH (parallel): D is NOT kept open -- only its price/verdict is needed",
      "dependencies": [],
      "instruction": "Find item D and report its price and whether it is under $60.",
      "role": "branch",
      "outputs": [{"name": "check_d__verdict", "description": "D price + under-$60 verdict + URL"}],
      "input_files": [],
      "init_from": null
    },
    {
      "id": "open_rest",
      "agent_type": "cua",
      "description": "SPINE TAIL: inherit A's heavy state (already open) and add only the cheap kept-open page B from its URL",
      "dependencies": ["assemble_a", "find_b"],
      "instruction": "A's pages are already open in this VM. Open product B's URL in a new tab and leave all tabs open.",
      "role": "spine",
      "outputs": [],
      "input_files": ["find_b__info"],
      "init_from": "assemble_a"
    }
  ],
  "aggregation": {
    "id": "final_aggregation",
    "agent_type": "manager",
    "description": "combine results into the final text answer",
    "dependencies": ["assemble_a", "find_b", "check_d"],
    "instruction": "Using A's and B's info and D's verdict, write the final summary for the user.",
    "role": "branch",
    "outputs": [],
    "input_files": ["assemble_a__info", "find_b__info", "check_d__verdict"],
    "init_from": null
  }
}

## Hard rules
- Subtask ids unique; `aggregation.id` is `"final_aggregation"`, `agent_type` `"manager"`, `init_from` null (it only reads artifacts).
- At least one CUA subtask; `subtasks` is never empty (the work always requires a CUA -- desktop/browser actions, never pure Q&A).
- ONLY `final_aggregation` has `agent_type: "manager"`. EVERY node in `subtasks` MUST have `agent_type: "cua"` -- including compile/plan/itinerary steps (they write a file in a VM, which is CUA work).
- Every `outputs` name is GLOBALLY UNIQUE across all nodes.
- Every `input_files` entry MUST equal some OTHER node's declared `outputs` name; a node never reads its own output.
- There is EXACTLY ONE spine: all `role: "spine"` nodes form a SINGLE `init_from` chain (one surviving deliverable VM). Never produce two unconnected operational nodes -- make all but the spine into branches that feed it artifacts. If any node lists this node as its `init_from`, this node MUST be `"spine"`. The aggregation node is `"branch"`.
- `init_from`, if set, MUST be one of the node's `dependencies` and reference a CUA subtask. (A `branch` node MAY have an `init_from`.)
- The producer of every artifact a node reads MUST be in that node's `dependencies`.
- Keep instructions specific and actionable; for a single short request one CUA can do, prefer the original task text verbatim as the instruction without over-prescribing steps.
\end{Verbatim}

\paragraph{User prompt.}
\begin{Verbatim}[breaklines=true, breaksymbolleft={}, fontsize=\scriptsize]
Here is the task to decompose into a spine-and-branch dependency graph:

<task>
{task}
</task>

If a screenshot of the current desktop is attached, use it to understand the initial VM state (open apps, visible files) and write more informed instructions.

Decompose the task (exploit only natural parallelism the task genuinely requires), then wire every node (outputs, input_files, init_from) per the rules. Output the JSON only.
\end{Verbatim}

\subsection{Step-budget prompting}
\label{app:prompt-sbp}

Step-budget prompting is introduced in Section~\ref{sec:method}: at every CUA step we append a short, escalating reminder to the agent's instruction, computed from the steps already taken and the hard step cap $M$. It adds no agents, tools, or extra model calls, only a few tokens per step and applies unchanged to any CUA, single- or multi-agent. Let $t$ be the 0-based index of the step about to be taken. The reminder is the header below followed by one of three tails selected by the used fraction $t/M$.

\paragraph{Header (always prepended).}
\begin{Verbatim}[breaklines=true, breaksymbolleft={}, fontsize=\scriptsize]
[STEP BUDGET] You are about to take step {t+1} of a HARD limit of {M} steps ({M-t} remaining). No further steps are possible after that.
\end{Verbatim}

\paragraph{Escalating tail (selected by $t/M$).}
\begin{Verbatim}[breaklines=true, breaksymbolleft={}, fontsize=\scriptsize]
# early, t/M < 0.5:
Work efficiently: avoid repeating actions or re-checking the same element, and commit to an answer as soon as you have enough information.

# past halfway, 0.5 <= t/M < 0.75:
You are past halfway. Prioritize finishing the deliverable (including saving any required file) over more exploration. If you already have enough information to answer, record it now rather than double-checking.

# near the cap, t/M >= 0.75:
You are running OUT of steps. Stop exploring immediately. If the task asks you to save a file, write your best current answer to that file NOW and then finish. Do not re-verify anything you already know and do not repeat any earlier action.
\end{Verbatim}

\clearpage
\section{The Odysseys-45 Development Subset}
\label{app:odysseys45}

Several ablations and diagnostics in this paper are run on a 45-task development
subset of Odysseys \citep{Jang2026-ip} rather than the full 200-task
benchmark, because each configuration must be executed in a live browser VM and then
scored by the rubric judge, which is costly to repeat across the many ablation cells.
This appendix documents how the subset was drawn and lists its task IDs so the split
is fully reproducible.

\paragraph{Sampling procedure.}
The full Odysseys benchmark contains 200 tasks whose difficulty labels are
skewed toward hard (45 easy, 46 medium, 109 hard). To obtain a compact subset that is
(i) balanced across difficulty and (ii) reproducible, we draw a \emph{stratified
random} sample of \textbf{15 tasks per difficulty level} (45 total). Sampling is without replacement and stratified only
by the difficulty label; no filtering by website, category, or rubric count is
applied. The resulting balance is 15\,/\,15\,/\,15 (easy\,/\,medium\,/\,hard), versus
the native 45\,/\,46\,/\,109. The subset spans 9 distinct starting websites (the majority begin from a Google search
portal, matching the full benchmark's entry distribution) and contains 264 atomic
rubric items in total (mean 5.9 per task, range 3--12).

\paragraph{Task IDs.}
The 45 selected task IDs, grouped by difficulty level:

\begingroup
\footnotesize\ttfamily\sloppy

\textbf{\upshape Easy (15).}\\
0106b570440ffe4427d5e916f39ec986ab3de917,
082aa17f3e88c3ce10796244e3677c5643dd19c9,\\
0ab48db6076089bbcf42047d162009a50eb9ca50,
0ce94d4e773eff1042a6920232f929a1da98c44d,\\
140960bb7293bdeeb6bcc60931681cb9b815351b,
156e2acc95361db4145f5bc313abb63807750089,\\
2504a7886c3dcb33f1aac7c5d2831985887e789e,
2cb0ed2a5df6053c6c982a5c5d436d25e006370f,\\
39255449e341c41a589b8a4e17f073be3a4809c9,
4246dec196c9a3382b4224c7ec3a34a20be9f43f,\\
69782bfcfdb3311496bc9048bf66915b33e692cd,
71f8e3e9b5a24f37f492fbf97b7d31e08e9a8d61,\\
795687ed918e45a6ad255215aa2a517b3e014aa5,
b1bd700090c23df9e9f6b7b9557ac418df602b8d,\\
b6b8ad71aa3112840790066d7d62b498babdfa5c.

\medskip
\textbf{\upshape Medium (15).}\\
041a4bee5d80a28567dc65bc2e41dd198672bfe2,
18ddad3e0781d4b8fb2e1998ff836a0b07d0cdce,\\
256342f13c0a03e080f92ee073153fe33a6881c0,
295f11f4eebda80a7551944fd9b6f4e01db92666,\\
53419597c0c8897d49f1af65f5255bf265edcfbf,
69f48a0950d532a2f04ff51abe4bf0e05ec5649e,\\
73c63095aeed43efb10a74eee7db7459c5ea9f84,
78ddd1aab59eebace5f6f523d90012aa6c871c54,\\
795bfe117e0f58e49ca37ae8e453a507859a2a2b,
8fcdeed84a0deb05342b07c26116792a5b6a6a3f,\\
946321e8a9788f485d360f619127a2e7b7e1693a,
b21a86441ddca8186175bfffcaae0358ed66eec4,\\
b4d11b2d7069bf45410b6784544504b23360b34a,
dd2eedbc88cb41cc69e43dd1da9de7255a81a966,\\
e4be2c73dc00107611cd648772a11fb15c18289b.

\medskip
\textbf{\upshape Hard (15).}\\
2a8418c2dccdaf5fe23ff143745cc5659d35fc69,
40735c71648c0ca0e6291d534685853acf1122c1,\\
54fdb6126b926e2a48b4e45ffc1fe303e873eb7f,
73f7a6bce89de66f106a669c85c7908331d3d1b7,\\
86cc69e23296a471c4e9e3da30d63ff54f31665f,
9acfcc050ae8de65ba5f5de787a47cef0589dd90,\\
9ad01a4a4bda2e8df7489c9831931b044c646a20,
a5724e1c94ac221f0a53765c51f625b7bc3cc58e,\\
a5d88b699a2524de24e157d9269d0c42a070edab,
a931f7baacfd7f1bcea8409bb8b3d84383734680,\\
d6007c19e6419c9eefdd57996fc151a2263b22fa,
e53065fe786881377e88667a80ccc2edcb321320,\\
ef766b69020befdc8e208f47401cb6bce5e9b931,
fb2f8bea3fa9528a581ce9e46bcc552c93e186a6,\\
fe1a5127a1329930e356744b7fd66a214592c630.

\endgroup

\clearpage
\section{CUA Action Space}
\label{app:actionspace}

Every worker is a Computer-Use
Agent (CUA) that operates the environment through the same low-level interface: at each step it receives a screenshot (plus its running action history) and emits a single tool call describing one GUI action, which is executed in the desktop VM via \texttt{pyautogui} before the next screenshot is taken. All task progress is made by pixel-level interaction. Coordinates are given as $(x,y)$ pixels; depending on configuration these are either absolute screen pixels or normalized to a $1000\times1000$ grid that the harness rescales to the true resolution.

Table~\ref{tab:actionspace} lists the complete action space, exposed to the open-weight backbones (\texttt{qwen3.6-27B}, \texttt{qwen3.7-plus}) as a single \texttt{computer\_use} tool with an \texttt{action} field and the arguments below. The proprietary \texttt{gpt-5.4-mini} backbone is driven through its provider's native computer-use tool, whose action set is equivalent and is translated to the identical \texttt{pyautogui} primitives in the same environment.
\begin{table}[t]
\centering\small
\caption{The CUA action space. Each step emits exactly one action. \texttt{coordinate}
is an $(x,y)$ pixel target. The \texttt{text} argument has two cases: (i) for
\texttt{type} and \texttt{answer} it is the literal string to type or return; (ii) for
click and scroll actions it is optional and names modifier key(s) to hold during the
action (e.g.\ \texttt{ctrl}, \texttt{shift}, \texttt{ctrl+shift}).}
\label{tab:actionspace}
\setlength{\tabcolsep}{5pt}
\begin{tabular}{l l p{0.46\linewidth}}
\toprule
\textbf{Action} & \textbf{Arguments} & \textbf{Description} \\
\midrule
\multicolumn{3}{l}{\textit{Mouse}} \\
\texttt{mouse\_move}       & \texttt{coordinate} & Move the cursor to $(x,y)$. \\
\texttt{left\_click}       & \texttt{coordinate}, [\texttt{text}] & Left-click at $(x,y)$; optional held modifier key(s). \\
\texttt{right\_click}      & \texttt{coordinate}, [\texttt{text}] & Right-click at $(x,y)$. \\
\texttt{middle\_click}     & \texttt{coordinate}, [\texttt{text}] & Middle-click at $(x,y)$. \\
\texttt{double\_click}     & \texttt{coordinate}, [\texttt{text}] & Double-click at $(x,y)$. \\
\texttt{triple\_click}     & \texttt{coordinate}, [\texttt{text}] & Triple-click at $(x,y)$ (executed as double-click). \\
\texttt{left\_click\_drag} & \texttt{coordinate} & Press, drag, and release to $(x,y)$. \\
\texttt{scroll}            & \texttt{pixels}, [\texttt{text}] & Vertical scroll by \texttt{pixels}; optional held modifier. \\
\texttt{hscroll}           & \texttt{pixels}, [\texttt{text}] & Horizontal scroll by \texttt{pixels}. \\
\addlinespace[2pt]
\multicolumn{3}{l}{\textit{Keyboard}} \\
\texttt{key}               & \texttt{keys} & Press the listed keys down in order then release in reverse (chords/hotkeys, e.g.\ \texttt{ctrl+c}). \\
\texttt{type}              & \texttt{text} & Type a string of text at the current focus. \\
\addlinespace[2pt]
\multicolumn{3}{l}{\textit{Control}} \\
\texttt{wait}              & \texttt{time} & Wait \texttt{time} seconds for the UI to settle, then re-observe. \\
\texttt{answer}            & \texttt{text} & Return a textual answer for an information-seeking task. \\
\texttt{terminate}         & \texttt{status} & End the task with \texttt{status}$\in\{$\texttt{success},\texttt{failure}$\}$ (\texttt{failure} = infeasible/unrecoverable). \\
\bottomrule
\end{tabular}
\end{table}
An episode ends when the agent emits
\texttt{terminate} (or \texttt{answer} for information-seeking tasks), or when the
per-worker step budget is exhausted. \texttt{terminate} with \texttt{status=failure} lets an agent declare a subtask infeasible rather than burning its remaining budget.
```latex
\clearpage
\section{Does a Weak Manager Degrade Spine-Branch?}
\label{sec:ablation-manager}

We swap only the manager and hold everything else fixed, comparing the default \texttt{claude-opus-4-8} with the substantially weaker \texttt{gpt-5.4-mini} across all three CUA backbones. Table~\ref{tab:ablation-manager} reports the macro score and success rate, with the Single-Agent result as reference.

\begin{table}[t]
\centering\small
\caption{Manager ablation for Spine-Branch on Odysseys-45: a strong
(\texttt{opus-4-8}) vs.\ a weak (\texttt{gpt-5.4-mini}) manager, with only the manager
changed.}
\label{tab:ablation-manager}
\setlength{\tabcolsep}{7pt}
\begin{tabular}{l l c c}
\toprule
\textbf{CUA backbone} & \textbf{Manager} & \textbf{Macro} & \textbf{SR} \\
\midrule
\texttt{Qwen3.7-Plus} & Single-Agent       & 0.830 & 0.622 \\
             & Spine-Branch, opus-4-8      & \textbf{0.906} & \textbf{0.733} \\
             & Spine-Branch, gpt-5.4-mini  & 0.819 & 0.644 \\
\addlinespace[2pt]
\texttt{Qwen3.6-27B}  & Single-Agent        & 0.585 & 0.311 \\
             & Spine-Branch, opus-4-8      & \textbf{0.751} &  \textbf{0.511} \\
             & Spine-Branch, gpt-5.4-mini  & 0.682 & 0.444 \\
\addlinespace[2pt]
\texttt{gpt-5.4-mini} & Single-Agent        & 0.596 & 0.444 \\
             & Spine-Branch, opus-4-8      & \textbf{0.818} & \textbf{0.689} \\
             & Spine-Branch, gpt-5.4-mini  & 0.678 & 0.556 \\
\bottomrule
\end{tabular}
\end{table}

\paragraph{Spine-Branch is moderately sensitive to manager quality, while its decomposition structure remains robust.}
Replacing \texttt{opus-4-8} with \texttt{gpt-5.4-mini} reduces the macro score by
$0.07$--$0.14$ across backbones, showing that manager strength matters. However, both
managers produce graphs with nearly identical structure ($\approx\!1.9$ spine and
$\approx\!3.5$ branch nodes per task), suggesting that the loss mainly stems from weaker per-node instructions and aggregation rather than degenerate decompositions. Importantly, even with the weak manager, Spine-Branch still outperforms the Single-Agent baseline on the two weaker executors (\texttt{Qwen3.6-27B}: $0.682$ vs.\
$0.585$; \texttt{gpt-5.4-mini}: $0.678$ vs.\ $0.596$). Only on the strongest executor,
\texttt{Qwen3.7-Plus}, does the gain disappear, where the strong single agent leaves less room for coordination gains. Overall, manager quality affects performance, but the Spine-Branch coordination scaffold remains effective across managers.
\clearpage
\section{Case Study}
\begin{figure*}[h]
\centering
\includegraphics[width=0.9\linewidth]{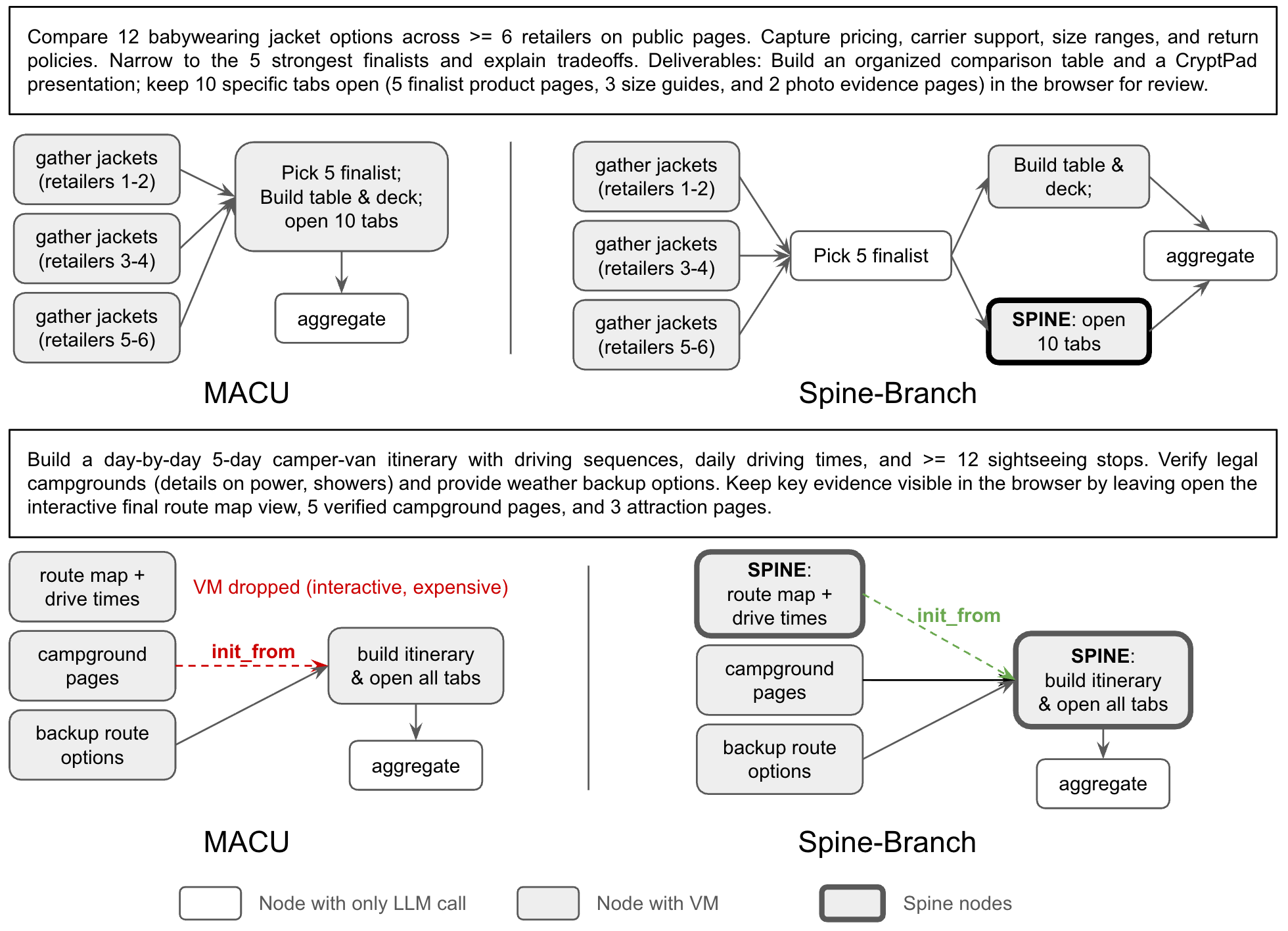}
\caption{Coordination graphs for the Babywearing-jacket task and Iceland Camper-Van Trip task. The graphs on the left are from MACU and the graphs on the right belong to Spine-Branch.}
\label{fig:case}
\end{figure*}
\begin{figure}[t]
\centering
\includegraphics[width=0.82\linewidth]{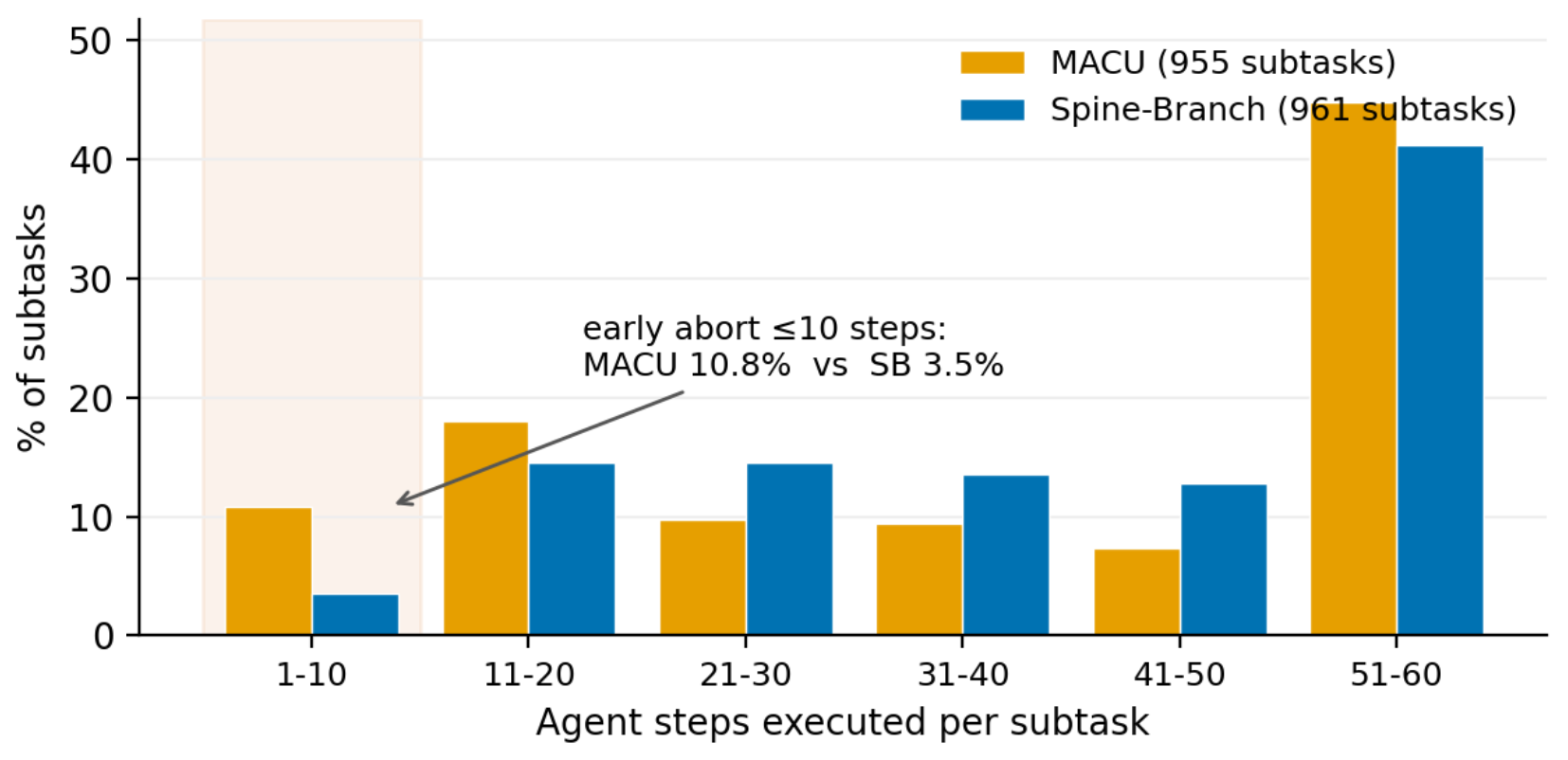}
\caption{\textbf{Agent step distribution of Spine-Branch and MACU on Odysseys with \texttt{Qwen3.6-27B} as backbone}.  MACU's manager continuously re-plans ($2.6$ replans/task) and
aggressively cancels underperforming subtasks: $\mathbf{10.8\%}$ of MACU subtasks abort
within $\le\!10$ steps, versus $\mathbf{3.5\%}$ for Spine-Branch (shaded bin). Because an
aborted subtask accrues almost no work, this pulls MACU's mean steps-per-subtask down to
$31.3$ (vs.\ $40.3$ for Spine-Branch) and, despite MACU \emph{attempting} more subtasks,
leaves it with fewer CUA tokens ($5.93$\,M vs.\ $6.69$\,M) and fewer agent actions ($250$
vs.\ $334$) per task. Spine-Branch fixes its plan up front and runs each subtask to
completion, so its heavier, uninterrupted execution costs more compute but yields far
higher completion (success rate $44.0\%$ vs.\ $28.5\%$; macro score $66.6\%$ vs.\
$53.9\%$)}
\label{fig:substep-hist}
\end{figure}

\end{document}